\documentclass[conference]{IEEEtran}
\IEEEoverridecommandlockouts
\usepackage{cite}
\usepackage{booktabs}
\usepackage{amsmath,amssymb,amsfonts}
\usepackage{siunitx}
\usepackage{algorithmic}
\usepackage{graphicx}
\usepackage{textcomp}
\usepackage{multirow}
\usepackage{xcolor}
\usepackage{array}
\usepackage{float}
\def\BibTeX{{\rm B\kern-.05em{\sc i\kern-.025em b}\kern-.08em
    T\kern-.1667em\lower.7ex\hbox{E}\kern-.125emX}}
\begin{document}

\title{Automated Construction of Medical Indicator Knowledge Graphs Using Retrieval Augmented Large Language Models}

\author{

\IEEEauthorblockN{Zhengda Wang}
\IEEEauthorblockA{The second hospital of Jilin University \\
Northeast Asia Active Aging Laboratory\\
Jilin, China\\
Email: wangzd9920@mails.jlu.edu.cn}

\and
\IEEEauthorblockN{Daqian Shi}
\IEEEauthorblockA{QMUL DERI \\UCL Institute Of Health Informatics, \\ Lonodn, UK \\
Email: d.shi@qmul.ac.uk}

\and
\IEEEauthorblockN{Jingyi Zhao}
\IEEEauthorblockA{The second hospital of Jilin University \\
Northeast Asia Active Aging Laboratory\\
Jilin, China\\
Email: zhaojingyi249@163.com}

\and
\IEEEauthorblockN{Xiaolei Diao}
\IEEEauthorblockA{
QMUL School of Electronic Engineering \\and Computer Science\\
London, UK\\
Email: xiaolei.diao1@gmail.com}

\and
\IEEEauthorblockN{Xiongfeng Tang}
\IEEEauthorblockA{The second hospital of Jilin University \\
College of Artificial Intelligence, \\Jilin University,
Jilin, China\\
Email: tangxf921@jlu.edu.cn}

\and
\IEEEauthorblockN{Yanguo Qin}
\IEEEauthorblockA{The second hospital of Jilin University \\
Northeast Asia Active Aging Laboratory\\
Jilin, China\\
Email: qinyg@jlu.edu.cn}}

\maketitle
\begin{abstract}
Artificial intelligence (AI) is reshaping modern healthcare by advancing disease diagnosis, treatment decision-making, and biomedical research. Among AI technologies, large language models (LLMs) have become especially impactful, enabling deep knowledge extraction and semantic reasoning from complex medical texts. However, effective clinical decision support requires knowledge in structured, interoperable formats. Knowledge graphs serve this role by integrating heterogeneous medical information into semantically consistent networks. Yet, current clinical knowledge graphs still depend heavily on manual curation and rule-based extraction, which is limited by the complexity and contextual ambiguity of medical guidelines and literature. To overcome these challenges, we propose an automated framework that combines retrieval-augmented generation (RAG) with LLMs to construct medical indicator knowledge graphs. The framework incorporates guideline-driven data acquisition, ontology-based schema design, and expert-in-the-loop validation to ensure scalability, accuracy, and clinical reliability. The resulting knowledge graphs can be integrated into intelligent diagnosis and question-answering systems, accelerating the development of AI-driven healthcare solutions.
\end{abstract}


\section{Introduction}
Artificial intelligence (AI) has become a central driver of innovation in modern medicine. Across domains such as medical imaging analysis, disease risk prediction, precision therapeutics, and personalized health management, AI has improved diagnostic accuracy, optimized treatment strategies, and enhanced healthcare delivery efficiency \cite{chandak2023building,al2024patient}. Within these technologies, large language models (LLMs) have introduced a paradigm shift in processing and understanding medical texts. Clinical guidelines, biomedical literature, and electronic health records are inherently complex, heterogeneous, and largely unstructured \cite{liang2024survey}. Traditional rule-based or dictionary-driven methods often fail to capture their semantic depth or adapt to evolving knowledge. By contrast, LLMs enable context-aware semantic reasoning, robust cross-domain generalization, and generative capabilities, making them powerful tools for medical knowledge extraction and utilization \cite{gao2025leveraging}.

However, the effective application of such knowledge requires structured, interoperable representations. Knowledge graphs serve as semantic scaffolds that transform fragmented information into interconnected networks of entities and relationships, supporting applications such as clinical decision support systems (CDSS), intelligent question answering, and biomedical research \cite{murali2023towards}. Yet, clinical knowledge graphs still rely heavily on manual curation and rule-based extraction to ensure factual precision and semantic consistency. This challenge stems from the nuanced and context-dependent nature of medical guidelines and scientific literature, where diagnostic criteria, therapeutic recommendations, and indicator definitions are embedded in complex narratives. Consequently, fully automated approaches remain uncommon, and existing methods struggle to accommodate the dynamic and cross-guideline characteristics of clinical indicators.

To address these limitations, we propose a framework that integrates retrieval-augmented generation (RAG) with LLMs for automated construction of medical indicator knowledge graphs. Semantic retrieval grounds generative outputs in authoritative sources, while LLM-based reasoning supports accurate entity and relation extraction. Additionally, an expert-in-the-loop (HITL) validation mechanism ensures clinical reliability while allowing iterative refinement \cite{wu2024medical}. This combination of automation, semantic rigor, and expert oversight provides a scalable and adaptable solution for building high-quality medical knowledge graphs, laying a robust foundation for intelligent healthcare systems.

\section{Related Work}

Recent advances in LLMs and biomedical knowledge graphs have driven significant progress in automating medical knowledge structuring and utilization. Gao et al. introduced MDKG, a contextualized mental disorder knowledge graph that addresses the lack of semantic context and indicator-level granularity in traditional KGs \cite{gao2025large}. Soman et al. developed a KG-optimized prompt generation framework, where structured biomedical knowledge informs LLM-based question answering \cite{soman2024biomedical}. In clinical applications, Yang et al. leveraged GPT-4 to construct a sepsis knowledge graph, marking the first integration of LLM reasoning into infectious disease modeling \cite{yang2025large}. To enhance factual precision, Wei et al. proposed MTL-KGV, a multitask verification framework that improves the reliability of automatically extracted biomedical triples \cite{wei2025biomedical}. Likewise, Xu et al. released PubMed Knowledge Graph 2.0, linking scientific publications, patents, and clinical trials into a scalable biomedical knowledge network \cite{xu2025pubmed}, while Wang et al. developed an end-to-end KG system supporting both construction and semantic querying across domains \cite{wang2025biomedical}.

Complementary research by Shi et al. has advanced semantic representation and reasoning in KGs. Their multidimensional KG framework structures entities via hierarchical semantic relations to enable adaptive path reasoning and personalized recommendation \cite{shi2020learning}. Building on this, Shi and Wang et al. introduced MRP2Rec, which captures high-order relational dependencies to improve interpretability and recommendation performance \cite{wang2020mrp2rec}. For multimodal consistency, their CharFormer model applies attention-guided feature fusion for image denoising while preserving structural semantics \cite{shi2022charformer}. Additionally, ZiNet, presented at ACL 2022, provides the first diachronic KG covering 3,000 years of Chinese linguistic evolution by integrating glyphs, radicals, and semantic hierarchies \cite{chi2022zinet}. Extending ZiNet, Diao et al. proposed a glyph-driven restoration network for Oracle Bone Inscriptions that leverages semantic priors to improve restoration under complex degradation \cite{diao2025oracle}.

Despite these advancements, existing methods still face limitations, including limited retrieval grounding, weak ontology integration, insufficient adaptability to multi-indicator clinical guidelines, and inadequate human–AI quality control. To address these issues, this study introduces a RAG-based pipeline combining guideline-grounded retrieval, ontology-driven schema design, and structured expert-in-the-loop validation, offering a scalable and semantically consistent solution for medical indicator knowledge graph construction.

\section{Methodology}

\subsection{Data Acquisition and Integration}

As shown in Fig. 1, the proposed framework integrates RAG with ontology-guided structuring to construct a medical knowledge graph. We first collect clinical practice guidelines from authoritative sources such as national health agencies, professional associations, and international healthcare organizations. These guidelines, often presented in varied formats, undergo a standardized preprocessing pipeline that includes content filtering, removal of non-informative elements, terminology normalization using controlled vocabularies, and unification of entity labels.

With expert collaboration, we define core clinical entity categories—including diseases, symptoms, diagnostic examinations, pharmacologic and surgical treatments, rehabilitation indicators, and postoperative metrics. This domain-specific schema supports subsequent ontology development and semantic extraction. Table 1 summarizes representative clinical indicator ranges across major systems, along with associated disease categories, including both directly and indirectly related conditions.

\begin{table*}[htbp]
\centering
\caption{Representative Clinical Indicators and Their Disease Associations}
\label{table:clinical_indicators}
\setlength{\tabcolsep}{2.5pt} 
\renewcommand{\arraystretch}{0.7}
\begin{tabular}{
  >{\raggedright\arraybackslash}p{1.2cm}
  >{\raggedright\arraybackslash}p{2.8cm}
  >{\raggedright\arraybackslash}p{2.8cm}
  >{\raggedright\arraybackslash}p{3.2cm}
  >{\raggedright\arraybackslash}p{3.4cm}
  >{\raggedright\arraybackslash}p{3.4cm}
}
\toprule
\textbf{System} & \textbf{Guideline} & \textbf{Indicator} & \textbf{Reference Range} & \textbf{Direct Disease} & \textbf{Indirect Diseases} \\
\midrule

\multirow{5}{*}[-10ex]{Endocrine}
  & American Thyroid Association 
  & Thyroid Stimulating Hormone 
  & $2$--$10$ mU/L 
  & Thyroid diseases 
  & Secondary thyroid diseases \\
  & American College of Physicians 
  & Testosterone 
  & Male: $300$--$1000$ ng/L\newline Female: $200$--$800$ ng/L 
  & Polycystic ovary syndrome 
  & Testicular dysgenesis \\
  & Chinese Society of Endocrinology 
  & Growth Hormone 
  & Children: $<20\,\mu$g/L\newline Male: $<2\,\mu$g/L\newline Female: $<10\,\mu$g/L 
  & Gigantism, acromegaly 
  & Pituitary dwarfism \\
  & Chinese Society of Endocrinology 
  & Human chorionic gonadotropin 
  & Male or non-pregnant female: $<5$ IU/L\newline Postmenopausal women: $<10$ IU/L 
  & Hydatidiform mole 
  & Elevated hCG in early pregnancy \\ 
  & Chinese Society of Cardiology 
  & Antidiuretic hormone 
  & $1.4$--$5.6$ pmol/L 
  & Nephrogenic diabetes insipidus 
  & Central diabetes insipidus \\
\midrule

\multirow{5}{*}[-10ex]{Circulatory}
  & World Health Organization 
  & Blood pressure 
  & $<120/80$ mmHg 
  & Hypertension, hypotension 
  & Cardiovascular diseases \\
  & American Heart Association 
  & Cholesterol 
  & $<200$ mg/dL 
  & Atherosclerosis 
  & Metabolic syndrome \\
  & Chinese College of Cardiovascular Physicians 
  & Creatine kinase 
  & Male: $50$--$310$ U/L\newline Female: $40$--$200$ U/L 
  & Atherosclerosis 
  & Myocarditis, rhabdomyolysis \\
  & European Society of Cardiology 
  & High-density lipoprotein (HDL) 
  & Male: $>40$ mg/dL\newline Female: $>50$ mg/dL 
  & Coronary heart disease 
  & Obesity, insulin resistance \\
  & European Society of Cardiology 
  & Low-density lipoprotein (LDL) 
  & $<100$ mg/dL 
  & Coronary heart disease 
  & Diabetic vascular complications \\
\midrule

\multirow{5}{*}[-10ex]{Urinary}
  & American College of Rheumatology 
  & Uric acid 
  & Male: $3.0$--$7.0$ mg/dL\newline Female: $2.5$--$6.5$ mg/dL 
  & Gout 
  & Chronic kidney disease \\
  & American Society of Nephrology 
  & Urinary red blood cells 
  & $<3$ per HPF 
  & Urolithiasis, glomerular disease 
  & Lupus nephritis, diabetic nephropathy \\
  & American Society of Nephrology 
  & Urinary white blood cells 
  & $<5$ per HPF 
  & Urinary tract infection 
  & Chronic renal insufficiency \\
  & Kidney Disease: Improving Global Outcomes 
  & Urinary protein 
  & 24~h: $<150$ mg 
  & Glomerular disease 
  & Hypertensive nephropathy \\
  & Kidney Disease: Improving Global Outcomes 
  & Glomerular filtration rate 
  & $90$--$120$ m\textsuperscript{2}/1.73 
  & Renal insufficiency, chronic kidney disease 
  & Cardiovascular diseases \\
\midrule

\multirow{5}{*}[-10ex]{Digestive}
  & World Gastroenterology Organisation 
  & Fecal occult blood test 
  & Negative 
  & Gastrointestinal bleeding 
  & Colorectal cancer \\
  & Chinese Society of Hepatology 
  & Transaminase 
  & $0$--$40$ U/L 
  & Hepatocellular injury 
  & Alcoholic liver disease \\
  & International Association of Pancreatology 
  & Lipase 
  & $13$--$60$ U/L 
  & Pancreatitis 
  & Renal insufficiency \\
  & American Cancer Society 
  & CA19-9 
  & $<37$ U/mL 
  & Pancreatic cancer 
  & Hepatobiliary diseases \\
  & American Cancer Society 
  & CEA 
  & $<5$ ng/mL 
  & Colorectal tumor 
  & Hepatic metastasis \\
\bottomrule
\end{tabular}
\end{table*}

\subsection{Ontology Design}

The ontology serves as the structural and semantic backbone of the medical knowledge graph, ensuring logical consistency, semantic alignment, and interpretability of the extracted knowledge. To develop a domain-specific and reusable ontology applicable across clinical scenarios, we combine LLM-based prompt engineering with iterative expert feedback. This enables both top-down schema definition and bottom-up refinement grounded in real guideline content. The resulting ontology includes core entity types such as diseases, diagnostic procedures, treatment strategies, medications, clinical indicators, and postoperative metrics. It also defines clinically meaningful relation types, including links between indications and treatment options, diagnostic procedures and threshold values, and postoperative indicators and follow-up plans.

In addition to entities and relationships, the ontology incorporates essential attributes such as disease prevalence, test frequencies, value ranges, risk classifications, and intervention thresholds. Hierarchical structures are embedded to enable nested classification of concepts, such as anatomical location within disease categories or the dependence of follow up indicators on treatment modalities. Logical constraints are applied to enforce semantic coherence and ensure completeness, such as requiring that every rehabilitation indicator be linked to a relevant clinical procedure. The resulting ontology is aligned with established biomedical standards, including SNOMED CT and UMLS, facilitating semantic interoperability and seamless integration with existing healthcare data systems and decision support platforms.

\subsection{Information Extraction}

To extract structured knowledge from clinical guidelines, we adopt a two-stage hybrid pipeline that integrates RAG with LLM inference. In the first stage, a semantic retrieval module performs document-level search using dense vector representations derived from pretrained biomedical embeddings. This enables efficient identification of contextually relevant guideline segments for specific extraction intents, offering greater robustness and relevance than rule-based or keyword-based matching.

In the second stage, the retrieved text is processed by an LLM to perform entity recognition, relation extraction, and attribute identification. The extracted results are organized into structured representations, such as subject–relation–object triples and attribute–value pairs, and aligned with the predefined ontology to ensure semantic and structural consistency. By combining targeted semantic retrieval with the flexible reasoning capabilities of LLMs, the framework achieves accurate, scalable, and semantically grounded knowledge graph construction from unstructured clinical guideline text.

\subsection{Knowledge Fusion and Graph Generation}

After information extraction, the structured knowledge elements are aligned with the predefined ontology to form a coherent and semantically consistent medical knowledge graph. This fusion process includes entity normalization to resolve synonyms and lexical variations, relation disambiguation to match extracted associations with ontology-defined semantics, and attribute integration to standardize numerical and categorical values.

To handle redundancy and conflict, duplicate or inconsistent entries are reconciled through rule-based prioritization and expert-guided resolution. These steps ensure semantic clarity, structural integrity, and clinical relevance in the final graph. Additionally, the framework supports interoperability with established biomedical ontologies such as UMLS and SNOMED CT, enhancing extensibility, reusability, and compatibility across diverse healthcare data environments and downstream applications.

\begin{figure}[htbp]
    \centering
    \includegraphics[width=0.75\columnwidth]{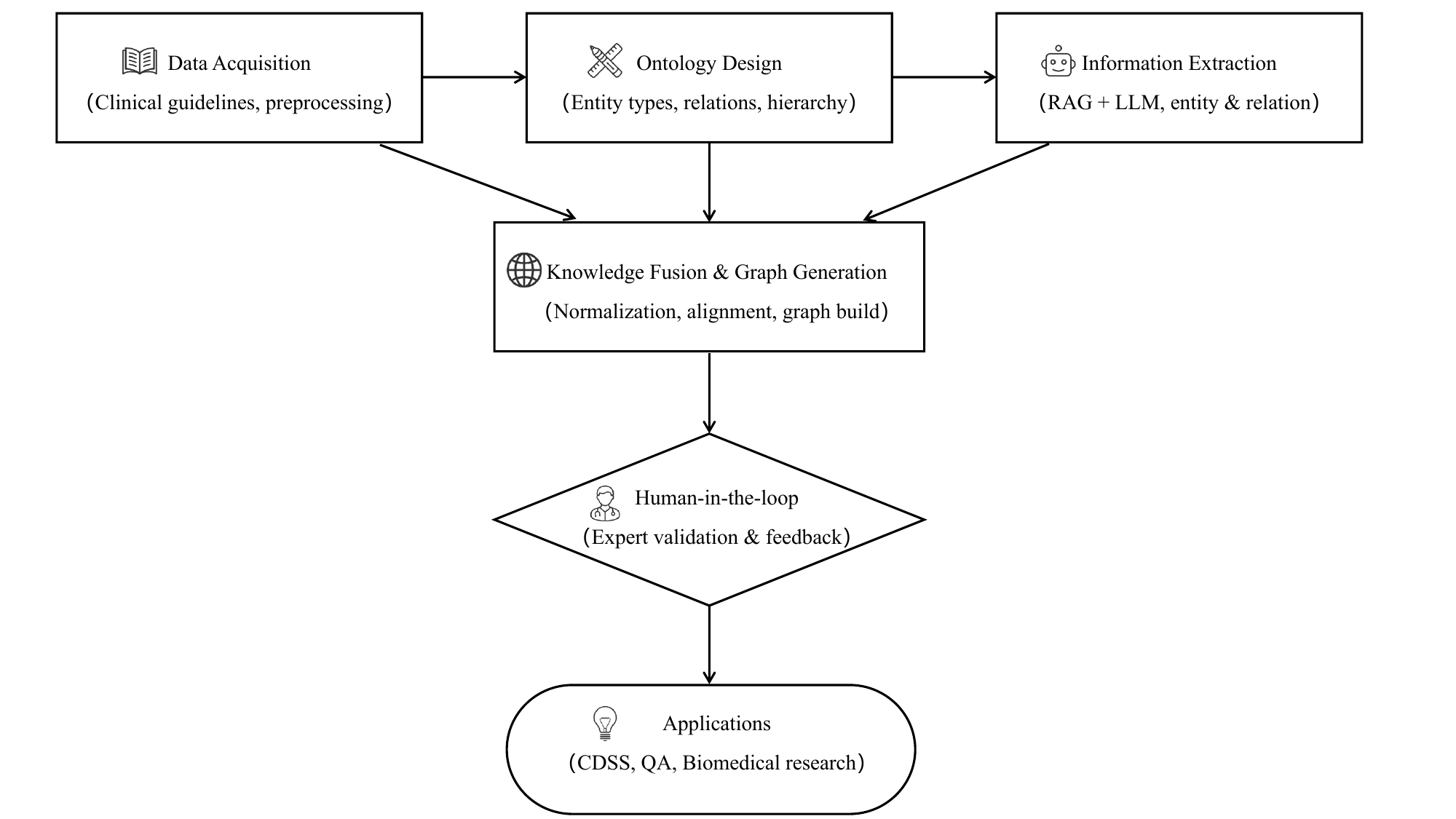}
    \caption{Workflow of the proposed knowledge graph construction framework.}
    \label{fig:framework}
\end{figure}

When inconsistencies, ambiguities, or missing information are identified during expert review, the feedback is reintegrated to refine prompt templates, adjust extraction rules, and improve LLM performance in subsequent cycles. This iterative human–AI collaboration forms a continuous quality control loop that safeguards reliability, semantic accuracy, and contextual relevance of the knowledge graph, while enabling ongoing optimization of the extraction pipeline and underlying knowledge representations.

\subsection{Applications}

The constructed medical knowledge graph supports a wide range of downstream applications in both clinical and research settings. In intelligent question-answering systems based on GraphRAG, it enables multi-hop reasoning, causal inference, and precise alignment with medical terminology, thereby improving both accuracy and interpretability. Within CDSS, it supports standardized diagnostic and treatment pathways, dynamic retrieval of guideline content, and personalized recommendations tailored to patient-specific contexts.

For medical research, the structured knowledge base accelerates hypothesis generation, enhances literature retrieval, and enables comparative analyses of clinical indicators across studies. By integrating RAG, ontology-driven schema modeling, and expert-in-the-loop validation, the proposed framework provides an efficient, scalable, and clinically meaningful approach for constructing high-quality medical knowledge graphs that align with the evolving needs of modern healthcare systems.

\section{Discussion}
The proposed framework represents a substantial advancement in medical knowledge graph construction by addressing the inherent limitations of traditional manual and rule-based approaches \cite{schafer2024biokgrapher}. Conventional methods are often labor-intensive, narrowly scoped, and lack the scalability needed to accommodate the rapid evolution of clinical knowledge \cite{yang2025retrieval}. In contrast, our method integrates RAG with LLMs to automate the extraction, normalization, and semantic integration of medical indicators from complex and unstructured guideline texts, building on recent advances in domain-specific generative model fine-tuning that have proven effective for clinical summarization and knowledge extraction \cite{wu2023knowlab}. The framework adopts a modular architecture consisting of data acquisition, ontology design, information extraction, and knowledge fusion, with each component independently optimizable to enhance overall performance \cite{shi2023recognizing}.

A key strength of the approach lies in the contextualized extraction enabled by semantic retrieval and ontology-guided structuring. Vector-based retrieval grounds LLM inference in clinically relevant source text, reducing hallucination and improving factual reliability, while the ontology provides a semantic scaffold that organizes extracted entities and relations into clinically meaningful hierarchies. To provide an initial quantitative check, we conducted an expert review of 240 extracted triples and confirmed 212 to be correct, resulting in an overall precision of 88 percent, demonstrating stable extraction performance at this stage. At present, the framework has standardized more than 120 clinical indicators derived from 38 authoritative guidelines spanning eight major physiological systems: musculoskeletal, respiratory, urinary, digestive, cardiovascular, endocrine, nervous, and immune–hematologic. Each indicator is linked to its guideline source, contextual definition, and associated disease entities, forming a continuously expanding and semantically aligned repository that supports cross-system comparison and shared biomarker interpretation.

To ensure scientific rigor and clinical reliability, a human-in-the-loop validation mechanism is embedded within the pipeline. Clinical experts conduct structured review, refine extraction prompts, and guide iterative improvement, forming a continuous quality-control loop that enhances knowledge fidelity and system adaptability. The resulting knowledge graphs are suitable for deployment in intelligent question-answering systems, clinical decision-support tools, and biomedical research platforms. While the framework demonstrates strong interoperability with established biomedical ontologies such as UMLS and SNOMED CT, future work will focus on automated graph updating, domain-specific LLM calibration, and maintaining a balanced integration of automation and expert oversight to support a continuously evolving and clinically reliable knowledge infrastructure. Sampled triples are validated with a concise checklist and escalated when needed, and expert feedback is used to iteratively refine prompts and extraction rules.

\section{Conclusion}
In this study, we developed an automated framework integrating RAG with LLMs to construct medical indicator knowledge graphs. The system has standardized over 120 indicators across eight physiological systems, demonstrating strong scalability and semantic consistency. Future work will focus on building a large-scale health visualization model that combines clinical guidelines with real-world hospital data to create personalized “health banks”, promoting precision and intelligent healthcare through data-driven insights. In addition, the framework will be expanded to support continuous knowledge updating, cross-domain ontology alignment, and multimodal data integration, paving the way toward an adaptive and interpretable medical intelligence ecosystem.

\section{Acknowledgment}
This work is supported by the Scientific Research Project of the Department of Education of Jilin Province (JJKH20250176KJ), and the Bethune Project of Jilin University (2025B37).

\bibliographystyle{IEEEtran}
\bibliography{ref}

@article{gao2025large,
  title={Large language model powered knowledge graph construction for mental health exploration},
  author={Gao, Shan and Yu, Kaixian and Yang, Yue and Yu, Sheng and Shi, Chenglong and Wang, Xueqin and Tang, Niansheng and Zhu, Hongtu},
  journal={Nature Communications},
  volume={16},
  number={1},
  pages={7526},
  year={2025},
  publisher={Nature Publishing Group UK London}
}

@article{soman2024biomedical,
  title={Biomedical knowledge graph-optimized prompt generation for large language models},
  author={Soman, Karthik and Rose, Peter W and Morris, John H and Akbas, Rabia E and Smith, Brett and Peetoom, Braian and Villouta-Reyes, Catalina and Cerono, Gabriel and Shi, Yongmei and Rizk-Jackson, Angela and others},
  journal={Bioinformatics},
  volume={40},
  number={9},
  pages={btae560},
  year={2024},
  publisher={Oxford University Press}
}

@article{yang2025large,
  title={Large Language Model--Driven Knowledge Graph Construction in Sepsis Care Using Multicenter Clinical Databases: Development and Usability Study},
  author={Yang, Hao and Li, Jiaxi and Zhang, Chi and Sierra, Alejandro Pazos and Shen, Bairong},
  journal={Journal of Medical Internet Research},
  volume={27},
  pages={e65537},
  year={2025},
  publisher={JMIR Publications Toronto, Canada}
}

@article{wei2025biomedical,
  title={Biomedical knowledge graph verification with multitask learning architectures},
  author={Wei, Chih-Ping and Tsai, Pei-Yuan and Li, Jih-Jane},
  journal={Journal of Biomedical Informatics},
  pages={104894},
  year={2025},
  publisher={Elsevier}
}

@article{xu2025pubmed,
  title={PubMed knowledge graph 2.0: Connecting papers, patents, and clinical trials in biomedical science},
  author={Xu, Jian and Yu, Chao and Xu, Jiawei and Torvik, Vetle I and Kang, Jaewoo and Sung, Mujeen and Song, Min and Bu, Yi and Ding, Ying},
  journal={Scientific Data},
  volume={12},
  number={1},
  pages={1018},
  year={2025},
  publisher={Nature Publishing Group UK London}
}

@article{wang2025biomedical,
  title={From biomedical knowledge graph construction to semantic querying: a comprehensive approach},
  author={Wang, Ling and Hao, Haoyu and Yan, Xue and Zhou, Tie Hua and Ryu, Keun Ho},
  journal={Scientific Reports},
  volume={15},
  number={1},
  pages={8523},
  year={2025},
  publisher={Nature Publishing Group UK London}
}

@article{chandak2023building,
  title={Building a knowledge graph to enable precision medicine},
  author={Chandak, Payal and Huang, Kexin and Zitnik, Marinka},
  journal={Scientific Data},
  volume={10},
  number={1},
  pages={67},
  year={2023},
  publisher={Nature Publishing Group UK London}
}

@article{al2024patient,
  title={Patient-centric knowledge graphs: a survey of current methods, challenges, and applications},
  author={Al Khatib, Hassan S and Neupane, Subash and Kumar Manchukonda, Harish and Golilarz, Noorbakhsh Amiri and Mittal, Sudip and Amirlatifi, Amin and Rahimi, Shahram},
  journal={Frontiers in Artificial Intelligence},
  volume={7},
  pages={1388479},
  year={2024},
  publisher={Frontiers Media SA}
}

@article{liang2024survey,
  title={A survey of LLM-augmented knowledge graph construction and application in complex product design},
  author={Liang, Xinxin and Wang, Zuoxu and Li, Mingrui and Yan, Zhijie},
  journal={Procedia CIRP},
  volume={128},
  pages={870--875},
  year={2024},
  publisher={Elsevier}
}

@article{murali2023towards,
  title={Towards electronic health record-based medical knowledge graph construction, completion, and applications: A literature study},
  author={Murali, Lino and Gopakumar, G and Viswanathan, Daleesha M and Nedungadi, Prema},
  journal={Journal of biomedical informatics},
  volume={143},
  pages={104403},
  year={2023},
  publisher={Elsevier}
}

@article{wu2024medical,
  title={Medical graph rag: Towards safe medical large language model via graph retrieval-augmented generation},
  author={Wu, Junde and Zhu, Jiayuan and Qi, Yunli and Chen, Jingkun and Xu, Min and Menolascina, Filippo and Grau, Vicente},
  journal={arXiv preprint arXiv:2408.04187},
  year={2024}
}

@article{gao2025leveraging,
  title={Leveraging medical knowledge graphs into large language models for diagnosis prediction: Design and application study},
  author={Gao, Yanjun and Li, Ruizhe and Croxford, Emma and Caskey, John and Patterson, Brian W and Churpek, Matthew and Miller, Timothy and Dligach, Dmitriy and Afshar, Majid},
  journal={Jmir Ai},
  volume={4},
  pages={e58670},
  year={2025},
  publisher={JMIR Publications Toronto, Canada}
}

@article{yang2025retrieval,
  title={Retrieval-augmented generation for generative artificial intelligence in health care},
  author={Yang, Rui and Ning, Yilin and Keppo, Emilia and Liu, Mingxuan and Hong, Chuan and Bitterman, Danielle S and Ong, Jasmine Chiat Ling and Ting, Daniel Shu Wei and Liu, Nan},
  journal={npj Health Systems},
  volume={2},
  number={1},
  pages={2},
  year={2025},
  publisher={Nature Publishing Group UK London}
}

@article{schafer2024biokgrapher,
  title={BioKGrapher: Initial evaluation of automated knowledge graph construction from biomedical literature},
  author={Sch{\"a}fer, Henning and Idrissi-Yaghir, Ahmad and Arzideh, Kamyar and Damm, Hendrik and Pakull, Tabea MG and Schmidt, Cynthia S and Bahn, Mikel and Lodde, Georg and Livingstone, Elisabeth and Schadendorf, Dirk and others},
  journal={Computational and Structural Biotechnology Journal},
  volume={24},
  pages={639--660},
  year={2024},
  publisher={Elsevier}
}

@article{shi2020learning,
  title={A learning path recommendation model based on a multidimensional knowledge graph framework for e-learning},
  author={Shi, Daqian and Wang, Ting and Xing, Hao and Xu, Hao},
  journal={Knowledge-Based Systems},
  volume={195},
  pages={105618},
  year={2020},
  publisher={Elsevier}
}

@article{wang2020mrp2rec,
  title={Mrp2rec: Exploring multiple-step relation path semantics for knowledge graph-based recommendations},
  author={Wang, Ting and Shi, Daqian and Wang, Zhaodan and Xu, Shuai and Xu, Hao},
  journal={IEEE Access},
  volume={8},
  pages={134817--134825},
  year={2020},
  publisher={IEEE}
}

@inproceedings{shi2022charformer,
  title={Charformer: A glyph fusion based attentive framework for high-precision character image denoising},
  author={Shi, Daqian and Diao, Xiaolei and Shi, Lida and Tang, Hao and Chi, Yang and Li, Chuntao and Xu, Hao},
  booktitle={Proceedings of the 30th ACM international conference on multimedia},
  pages={1147--1155},
  year={2022}
}

@inproceedings{chi2022zinet,
  title={ZiNet: Linking Chinese characters spanning three thousand years},
  author={Chi, Yang and Giunchiglia, Fausto and Shi, Daqian and Diao, Xiaolei and Li, Chuntao and Xu, Hao},
  booktitle={Findings of the association for computational linguistics: ACL 2022},
  pages={3061--3070},
  year={2022}
}

@article{diao2025oracle,
  title={Oracle bone inscription image restoration via glyph extraction},
  author={Diao, Xiaolei and Shi, Daqian and Cao, Wei and Wang, Ting and Qi, Ruihua and Li, Chuntao and Xu, Hao},
  journal={npj Heritage Science},
  volume={13},
  number={1},
  pages={321},
  year={2025},
  publisher={Springer International Publishing Cham}
}

@inproceedings{shi2023recognizing,
  title={Recognizing entity types via properties},
  author={Shi, Daqian and Giunchiglia, Fausto},
  booktitle={Formal Ontology in Information Systems},
  pages={195--209},
  year={2023},
  organization={IOS Press}
}

@inproceedings{wu2023knowlab,
  title={KnowLab at RadSum23: comparing pre-trained language models in radiology report summarization},
  author={Wu, Jinge and Shi, Daqian and Hasan, Abul and Wu, Honghan},
  booktitle={Proceedings of the Annual Meeting of the Association for Computational Linguistics},
  pages={535--540},
  year={2023},
  organization={ACL}
}

\end{document}